\documentclass[11pt]{article}
\usepackage[margin=1in]{geometry}
\usepackage[T1]{fontenc}
\usepackage[utf8]{inputenc}
\usepackage{amsmath,amssymb}
\usepackage{graphicx}
\usepackage{booktabs}
\usepackage{array}
\usepackage{float}
\usepackage{hyperref}
\title{MedSkillAudit: A Domain-Specific Audit Framework for Medical Research Agent Skills}
\author{}
\date{}
\begin{document}
\pagestyle{empty}
\maketitle

\begin{center}
{\large
Yingyong Hou$^{2*}$ \quad
Xinyuan Lao$^{1*}$ \quad
Huimei Wang$^{1,2*}$ \quad
Qianyu Yao$^{1,2\dagger}$ \quad
Wei Chen$^{1,2\dagger}$ \\
Bocheng Huang$^{1,2\dagger}$ \quad
Fei Sun$^{1,2\ddagger}$ \quad
Yuxian Lv$^{1,2\ddagger}$ \quad
Weiqi Lei$^{1,2\ddagger}$ \quad
Xueqian Wen$^{1,2}$ \\
Shengyang Xie$^{1,2}$ \quad
Pengfei Xia$^{1,2}$ \quad
Zhujun Tan$^{1,2}$
}

{\normalsize
$^{\dagger}$Equal first contribution \quad
$^{\ddagger}$Equal second contribution \quad
$^{*}$Co-corresponding authors
}

\vspace{0.5em}

{\small
$^{1}$AIPOCH PTE. LTD., Singapore \\
$^{2}$Department of Pathology, Zhongshan Hospital, Fudan University, Shanghai, China \\
}

\vspace{0.5em}

\end{center}

\begin{abstract}
\noindent
\textbf{Background:} Agent skills are increasingly deployed as modular, reusable capability units in AI agent systems. Medical research agent skills require safeguards beyond general-purpose evaluation, including scientific integrity, methodological validity, reproducibility, and boundary safety. To develop and preliminarily evaluate a domain-specific audit framework for medical research agent skills, with a focus on reliability against expert review. \textbf{Methods:} We developed MedSkillAudit (skill-auditor@1.0), a layered framework assessing skill release readiness before deployment. We evaluated 75 skills across five medical research categories (15 per category), independently reviewed by two experts who assigned a quality score (0--100), an ordinal release disposition (Production Ready / Limited Release / Beta Only / Reject), and a high-risk failure flag. System--expert agreement was quantified using ICC(2,1) and linearly weighted Cohen's $\kappa$, benchmarked against the human inter-rater baseline. \textbf{Results:} The mean consensus quality score was 72.4 (SD = 13.0); 57.3\% of skills fell below the Limited Release threshold. MedSkillAudit achieved ICC(2,1) = 0.449 (95\% CI: 0.250--0.610), exceeding the human inter-rater ICC of 0.300. System--consensus score divergence (SD = 9.5) was smaller than inter-expert divergence (SD = 12.4), with no directional bias (Wilcoxon p = 0.613). Protocol Design showed the strongest category-level agreement (ICC = 0.551); Academic Writing showed a negative ICC (-0.567), reflecting a structural rubric--expert mismatch. \textbf{Conclusions:} Domain-specific pre-deployment audit may provide a practical foundation for governing medical research agent skills, complementing general-purpose quality checks with structured audit workflows tailored to scientific use cases.
\end{abstract}

\noindent\textbf{Keywords:} medical AI evaluation; agent skills; pre-deployment governance; reliability study; automated quality audit; intraclass correlation

\begin{center}
\includegraphics[width=0.92\textwidth]{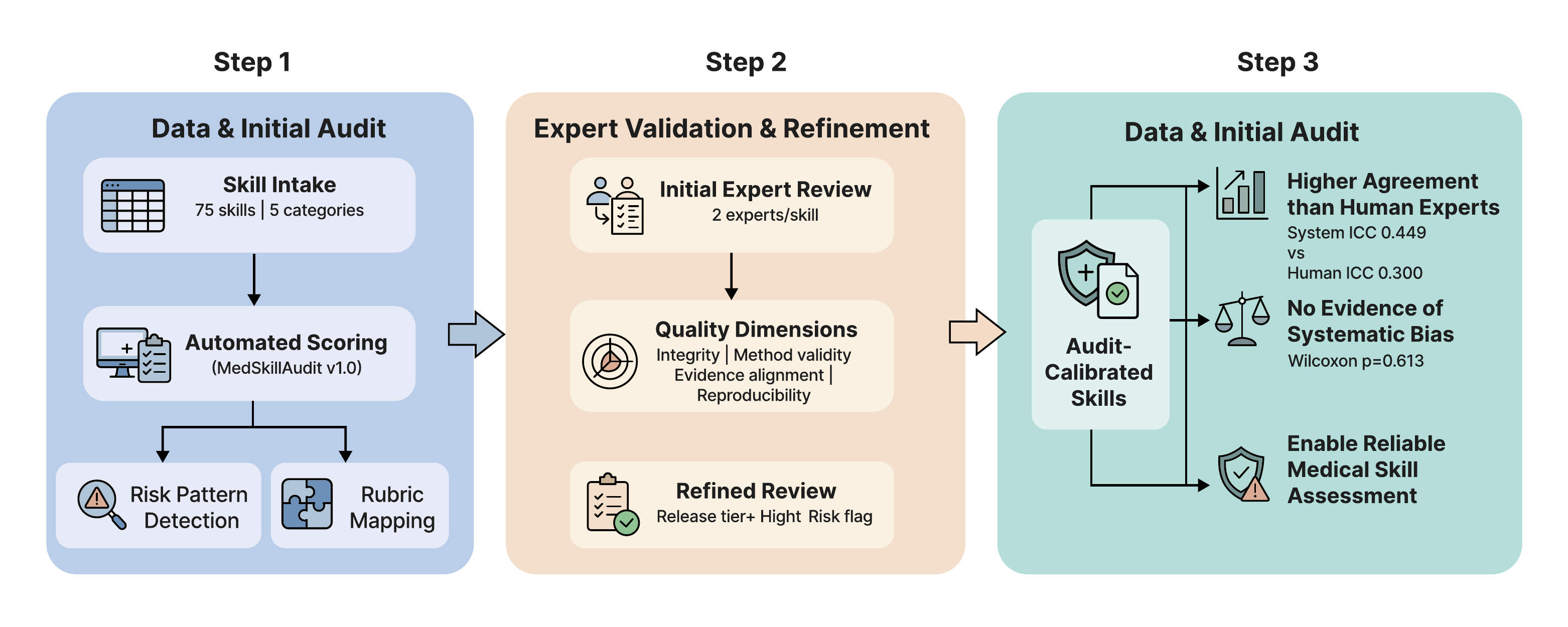}

{\small\textbf{Graphical abstract.} MedSkillAudit framework, evaluation workflow, and principal findings.}
\end{center}

\section{Introduction}

AI agent systems are increasingly being extended through skills: modular, reusable packages that encapsulate task-specific instructions, procedural guidance, and, in some cases, executable resources [1, 2]. Compared with one-off prompts, skill packages are more structured, portable, and auditable, allowing capabilities to be reused across workflows, versioned over time, and evaluated as independent artifacts [2]. As skill-based agent ecosystems expand, the quality and safety of the skills themselves become increasingly important.

Recent work has begun to treat skills as first-class objects of study. One line of research focuses on skill utility, asking whether skills improve downstream agent performance on benchmark tasks [1]. Another line focuses on skill quality, examining whether skill artifacts are safe, complete, executable, maintainable, and otherwise suitable for reuse [2]. Together, these studies establish that skills are not merely stylistic prompt wrappers, but operational units that can shape agent behavior in measurable ways [1, 2].

However, medical research agent skills introduce additional requirements not fully addressed by general-purpose evaluation. Prior work on medical large language models and biomedical agents has shown that strong apparent performance does not eliminate concerns about reliability, calibration, safety, tool use, and domain-specific evaluation in high-stakes settings [4, 6--11]. A skill may appear structurally complete yet remain scientifically unreliable---producing unsupported claims, misaligned analytical choices, or irreproducible guidance that might influence research reasoning rather than merely degrade surface-level output [3, 5, 12--15]. These concerns underscore the need for domain-specific audit frameworks that evaluate not only structural and functional quality but also scientific integrity and release readiness. Furthermore, existing evaluation approaches tend to emphasize ranking or filtering rather than iterative improvement. Given that skill packages are explicitly documented and designed for reuse [1, 2], effective audit systems should also provide actionable feedback to enhance methodological rigor, safety, and suitability for responsible deployment.

Existing evaluation frameworks for medical AI systems fall into three broad categories. Benchmark-based capability assessments (e.g., USMLE performance [6, 7], expert-level question answering [8]) measure what a model can do on standardized test items, but do not address whether a skill artifact is safe, reproducible, and suitable for deployment as a reusable research component. Agentic evaluation environments (e.g., MedAgentBench [11]) assess task completion in simulated clinical settings, but evaluate emergent agent behavior rather than the auditable properties of packaged skill artifacts. General-purpose code quality tools apply software engineering criteria that do not account for scientific computing semantics, domain-specific safety requirements, or the distinction between runtime crashes and methodologically incorrect outputs. MedSkillAudit addresses a distinct layer of the evaluation stack: pre-deployment governance of the skill artifact itself, evaluated against criteria derived from scientific standards and deployment risk rather than downstream task performance.

In this study, we present MedSkillAudit, a domain-specific audit framework for medical research agent skills. Rather than claiming a large-scale benchmark, we focus on a foundational question relevant to pre-deployment governance: whether a domain-specific audit framework can evaluate medical research agent skills with meaningful reliability compared with expert review. To investigate this question, we curated an evaluation set of 75 skills spanning five medical research-related categories and conducted a reliability study (Experiment 1) comparing framework outputs with independent expert review.

This work makes three contributions. First, we propose a layered audit framework tailored to the specific risks and requirements of medical research agent skills. Second, we provide an initial reliability evaluation on a 75-skill corpus by comparing framework outputs with independent expert review. Third, we show preliminary evidence that system--expert agreement exceeds the human inter-rater baseline in terms of divergence magnitude, suggesting that structured automated audit may be a viable complement to human evaluation in pre-deployment governance workflows for medical research agent skills.

\section{Methods}

\subsection{Study Design}

We developed MedSkillAudit, a domain-specific audit framework for evaluating the release readiness of medical research agent skills before deployment. The framework was designed for reusable skill artifacts organized around a central SKILL.md specification document (a structured Markdown file defining the skill's name, description, input/output schema, and execution instructions) and optionally accompanied by executable scripts, templates, or external API integrations [1, 2]. In this study, skills were treated as standalone evaluation objects rather than isolated outputs from a single prompting instance.

This study addressed a single primary research question: whether MedSkillAudit can generate evaluations that align meaningfully with expert review. Accordingly, the study consisted of a reliability study, performed on the full evaluation set of 75 skills, in which system outputs were compared against independent expert review using standard agreement statistics.

\subsection{Skill Evaluation Set}

We constructed an evaluation set of 75 medical research-related skills, with 15 skills sampled from each of five functional categories (Table 1). Skills were drawn from four successive development cycles produced by two independent research teams, with random sampling applied within each category to ensure coverage of both earlier and more mature iterations. This sampling strategy was chosen to capture a realistic range of skill quality at the pre-deployment stage, rather than to construct an optimized showcase corpus.

The five categories were selected to cover a range of common medical research workflows: (1) \textit{Evidence Insight}, encompassing literature retrieval, appraisal, and synthesis; (2) \textit{Protocol Design}, covering experimental design generation and statistical planning; (3) \textit{Data Analysis}, including computational analysis and bioinformatics code generation; (4) \textit{Academic Writing}, comprising scientific manuscript and document generation; and (5) \textit{Other}, a general utility category for skills not fitting the preceding four.

Each skill was treated as a reusable artifact intended to support a class of research tasks rather than solve a single instance. For each skill, we recorded metadata including category, estimated task complexity (Simple / Moderate / Complex, determined programmatically based on reference file count, SKILL.md length, and task branching depth), and execution mode (Mode A: prompt-only; Mode B: CLI/script-based; Mode D: hybrid script and API). The evaluation set comprised 22 Mode A skills (29.3\%), 42 Mode B skills (56.0\%), and 11 Mode D skills (14.7\%).

\begin{table}[H]
\centering
\caption{Evaluation set composition.}
\scriptsize
\resizebox{\textwidth}{!}{%
\begin{tabular}{p{0.34\textwidth} p{0.11\textwidth} p{0.35\textwidth}}
\toprule
Category & n & Execution Mode (A / B / D) \\
\midrule
Evidence Insight & 15 & 2 / 12 / 1 \\
Protocol Design & 15 & 12 / 2 / 1 \\
Data Analysis & 15 & 0 / 14 / 1 \\
Academic Writing & 15 & 8 / 7 / 0 \\
Other & 15 & 0 / 7 / 8 \\
Total & 75 & 22 / 42 / 11 \\
\bottomrule
\end{tabular}%
}
\end{table}

\subsection{MedSkillAudit Framework}

This study evaluated MedSkillAudit version 1.0 (skill-auditor@1.0). Based on findings from Experiment 1, the framework was iteratively refined to version 1.1.0 after data collection was complete; the rationale and specific changes are described in Section 4.3. These post-hoc refinements have not yet been evaluated in a controlled experiment. MedSkillAudit is a layered audit pipeline that combines an automated Python pre-screening script (Steps 1--3) with a Claude-driven evaluation agent (Steps 4--8). The pipeline evaluates whether a medical research skill is suitable for release and reuse, and produces both an ordinal release disposition and structured feedback for revision. An overview of the framework architecture is presented in Figure 1. The framework is grounded in the ISO/IEC 25010 software quality model [16], the OpenSSF security framework, Shneiderman's usability principles, and domain-specific medical research quality standards.

\begin{figure}[H]
\centering
\includegraphics[width=0.92\textwidth]{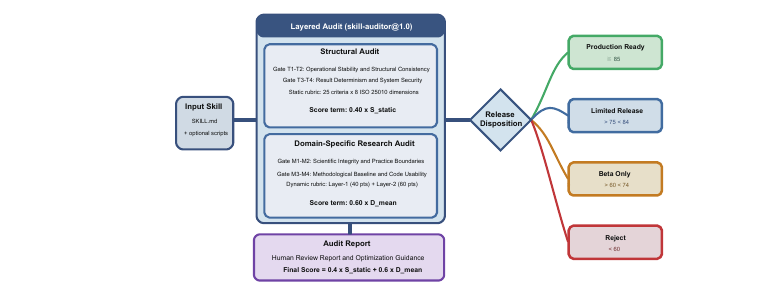}
\caption{Overview of the MedSkillAudit framework (skill-auditor@1.0). Input Skill Artifacts (SKILL.md and optional scripts) enter a layered audit pipeline. The Structural Audit layer applies Veto Gate 1 (T1 Operational Stability, T2 Structural Consistency, T3 Result Determinism, T4 System Security) and computes a static quality score ($S_{\mathrm{static}}$) across 25 criteria and 8 ISO 25010-aligned dimensions. The Domain-Specific Research Audit layer applies Veto Gate 2 (M1 Scientific Integrity, M2 Practice Boundaries, M3 Methodological Baseline, M4 Code Usability) after dynamic execution testing, computing a mean dynamic score ($\bar{D}$) via a two-layer rubric (Layer 1 generic, 40 pts; Layer 2 category-specific, 60 pts). Any veto failure results in immediate Reject regardless of score. The final quality score is: Final Score = $0.4 \times S_{\mathrm{static}} + 0.6 \times \bar{D}$. Release disposition is assigned to one of four tiers: Production Ready ($\geq$85), Limited Release (75--84), Beta Only (60--74), or Reject ($<60$). The Audit Report (Steps 7--8) outputs structured JSON and Markdown containing per-dimension scores, veto evidence, and optimization guidance for skill improvement.}
\end{figure}

\subsubsection{Structural Audit (Veto Gate 1)}

The first layer evaluates the structural integrity of the skill artifact via four hard-gate dimensions: \textit{Operational Stability} (T1; crash rate $\leq$ 20\%, no unresolvable dependency conflicts), \textit{Structural Consistency} (T2; compliant SKILL.md schema with mandatory \texttt{name} and \texttt{description} fields, internally consistent return types), \textit{Result Determinism} (T3; no unseeded random number calls, no unbounded loops), and \textit{System Security} (T4; no unsanitized \texttt{eval}/\texttt{exec} calls, no prompt injection vectors). Any dimension receiving a FAIL verdict triggers immediate rejection; the remaining evaluation steps are not executed.

\subsubsection{Domain-Specific Research Audit (Veto Gate 2)}

The second veto gate is applied after dynamic output evaluation, exclusively for categories 1--4. It checks four scientific integrity dimensions: \textit{Scientific Integrity} (M1; no fabricated citations, DOIs, sample sizes, or p-values), \textit{Practice Boundaries} (M2; no direct diagnostic conclusions, required medical disclaimers present), \textit{Methodological Baseline} (M3; no logical fallacies such as conflating correlation with causation), and \textit{Code Usability} (M4; no syntax errors or missing core dependencies in generated code; marked N/A for categories 1 and 4 when no code is produced). Any FAIL triggers rejection regardless of numeric score.

\subsubsection{Scoring and Release Disposition}

Skills passing both veto gates received a final quality score:

$$\text\ Final Score\ = 0.4 \times S\_\{\text\{static\}\} + 0.6 \times \bar\{D\}$$

where $S_{\text{static}}$ is the static quality score (0--100) aggregating 25 criteria across 8 ISO 25010-aligned dimensions evaluated against the skill's specification and code, and $\bar{D}$ is the mean dynamic score across $N$ test inputs (3, 5, or 7, scaled to assessed complexity). Dynamic scoring applies a two-layer rubric per output: Layer 1 evaluates generic output quality across functional correctness, reliability, efficiency, and scope adherence (40 points); Layer 2 applies a category-specific specialized rubric (60 points) covering domain-relevant dimensions such as search strategy rigor (Category 1), design soundness (Category 2), code executability (Category 3), terminology precision (Category 4), and task completion (Category 5). Boolean assertion checks evaluate structural completeness per output but are not incorporated into the numeric score.

Release disposition was assigned according to fixed score thresholds: $\geq$ 85 (\textit{Production Ready}\textit{), 75--84 (}\textit{Limited Release}\textit{), 60--74 (}\textit{Beta Only}\textit{), < 60 (}\textit{Reject}). A veto failure overrides the numeric grade and results in Reject regardless of score.

\subsection{Expert Review Protocol}

Each of the 75 skills was independently reviewed by two expert evaluators (Expert 1, Expert 2) with relevant medical research background, using the same rubric dimensions as the MedSkillAudit framework. Experts interacted with each skill in a standardized evaluation environment, executing representative tasks and reviewing skill outputs. For every skill, each expert assigned: (1) an overall quality score on a continuous 0--100 scale; (2) an ordinal release disposition using the same four-level scale as the automated system; and (3) a binary high-risk flag (Y/N) to indicate whether observed failures posed patient safety or scientific integrity risks warranting non-release treatment.

Expert 1 evaluated skills S001--S045 using individual structured rating files, with skills S046--S075 evaluated by a second rater on the same team using an identical rubric. Expert 2 evaluated skills S001--S045 via structured summary workbooks and skills S046--S075 via individual files. All ratings were recorded in standardized spreadsheet templates. A cross-reference verification pass confirmed zero score discrepancies and zero disposition rank discrepancies between recorded ratings and the primary database. Expert 1 and Expert 2 used different rating formats (individual files vs. summary workbooks) to minimize shared method variance, though this design precludes standard paired analysis and is acknowledged as a limitation.

\subsection{Consensus Derivation}

The expert consensus quality score was computed as the arithmetic mean of Expert 1 and Expert 2 scores. One exception applies: for S010, Expert 1 assigned no numeric score (all four Structural Veto dimensions failed; the skill contained no executable code, rendering quality scoring uninformative), and the consensus quality score was set to the Expert 2 score alone (59.6). Consensus disposition was derived by adjudication when experts differed: for one-rank disagreements, the disposition closer to the score-weighted mean was adopted; for larger disagreements, the more conservative (lower-release) disposition was used. Adjudication was flagged as required for all cases of expert rank disagreement. Consensus high-risk flag was set to Y (both agreed Y), N (both agreed N), or Unclear (disagreement).

\subsection{Statistical Analysis}

All analyses were performed in Python 3.9 using pandas [17], pingouin [18], scipy [19], and scikit-learn [20].

Inter-rater score agreement was quantified using the intraclass correlation coefficient ICC(2,1) --- two-way random effects, single measures, absolute agreement --- following the model selection guidelines of Koo and Li [21]. Disposition-rank agreement was quantified using linearly weighted Cohen's $\kappa$ [22], which assigns partial credit proportional to rank distance and is appropriate for ordinal four-category scales. We additionally computed the distribution of absolute score differences $|s_{E1} - s_{E2}|$ to characterize the human rater divergence baseline, which served as a reference for interpreting the magnitude of system--consensus divergence. System--consensus agreement used the same ICC(2,1) and weighted $\kappa$ statistics. Systematic directional bias was tested with the Wilcoxon signed-rank test (two-sided). Agreement was visualized using a Bland-Altman plot [23], with limits of agreement defined as $\bar{\Delta} \pm 1.96 \cdot SD_\Delta$ where $\Delta = s_\text{sys} - s_\text{con}$. Analyses stratified by skill category were treated as descriptive rather than confirmatory, given the per-category sample sizes of 15.

Skills were flagged for optimization if they met any of the following pre-specified criteria: (1) consensus Reject disposition; (2) Beta Only disposition with score < 65; (3) expert adjudication required; (4) system--consensus rank gap $\geq$ 2; or (5) high-risk flag Y or Unclear.

\section{Results}

\subsection{Baseline Quality Assessment}

The 75 skills yielded a mean consensus quality score of 72.4 (SD = 13.0; median = 73.2; IQR = 64.4--84.1; range = 40.0--90.8). By release disposition, 17 skills (22.7\%) were Production Ready, 15 (20.0\%) Limited Release, 31 (41.3\%) Beta Only, and 12 (16.0\%) Reject (Figure 2). The modal outcome was Beta Only, and 57.3\% of skills fell below the Limited Release threshold, indicating that the majority of skills were not deployment-ready at baseline.

\begin{figure}[H]
\centering
\includegraphics[width=0.92\textwidth]{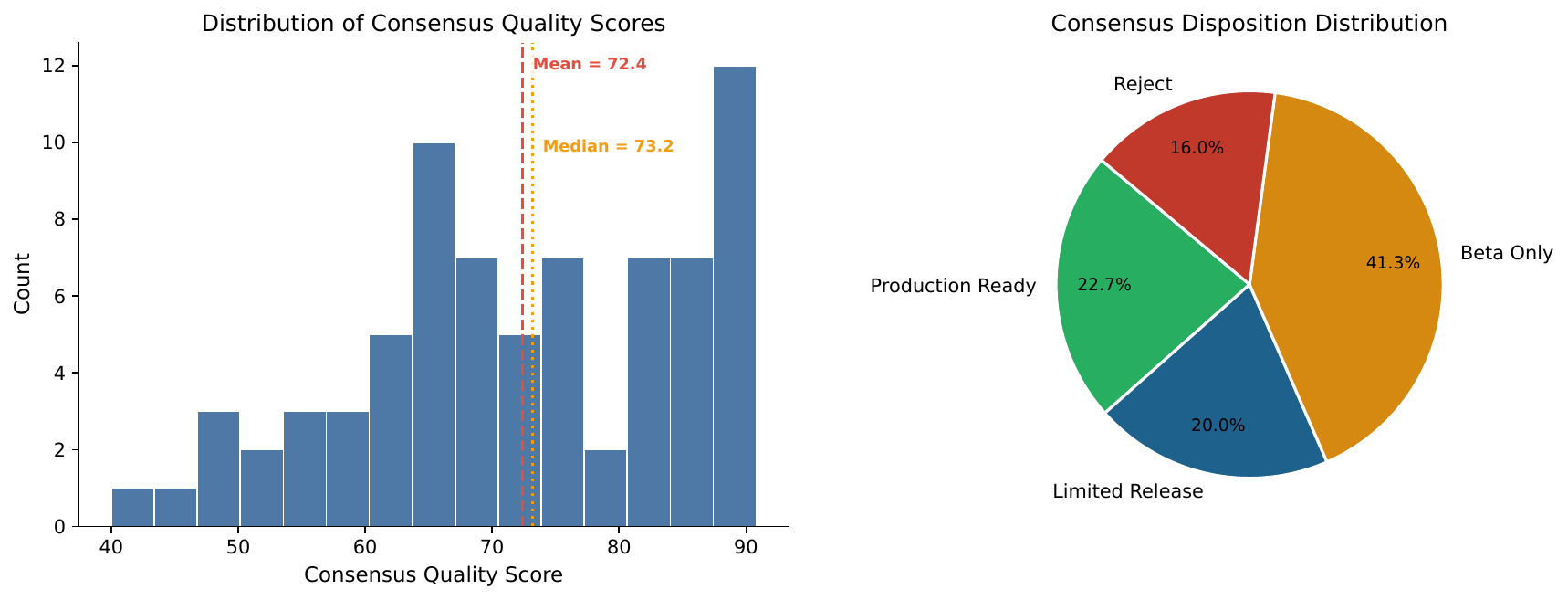}
\caption{Overall quality score distribution and release disposition. Left: histogram of consensus quality scores (n = 75) with disposition thresholds shown as vertical dashed lines. Right: pie chart showing the proportion of skills in each release disposition tier.}
\end{figure}

Marked quality variation was observed across categories (Figure 3, Table 2). Protocol Design achieved the highest mean consensus score (86.2 $\pm$ 3.8) and the most compressed score distribution (range: 80.0--90.7). Academic Writing had the lowest mean score (62.7 $\pm$ 7.2), with 5 of 15 skills receiving a Reject disposition. Data Analysis exhibited the widest score variance (70.7 $\pm$ 15.3), driven by a subgroup of skills with dependency-related runtime failures. Prompt-only skills (Mode A; n = 22) achieved a higher mean consensus score (77.9 $\pm$ 12.9) than script-based skills (Mode B: 70.1 $\pm$ 13.0; Mode D: 70.2 $\pm$ 10.4), consistent with the additional failure vectors introduced by dependency management and runtime execution in code-based skills (Figure 4).

\begin{figure}[H]
\centering
\includegraphics[width=0.92\textwidth]{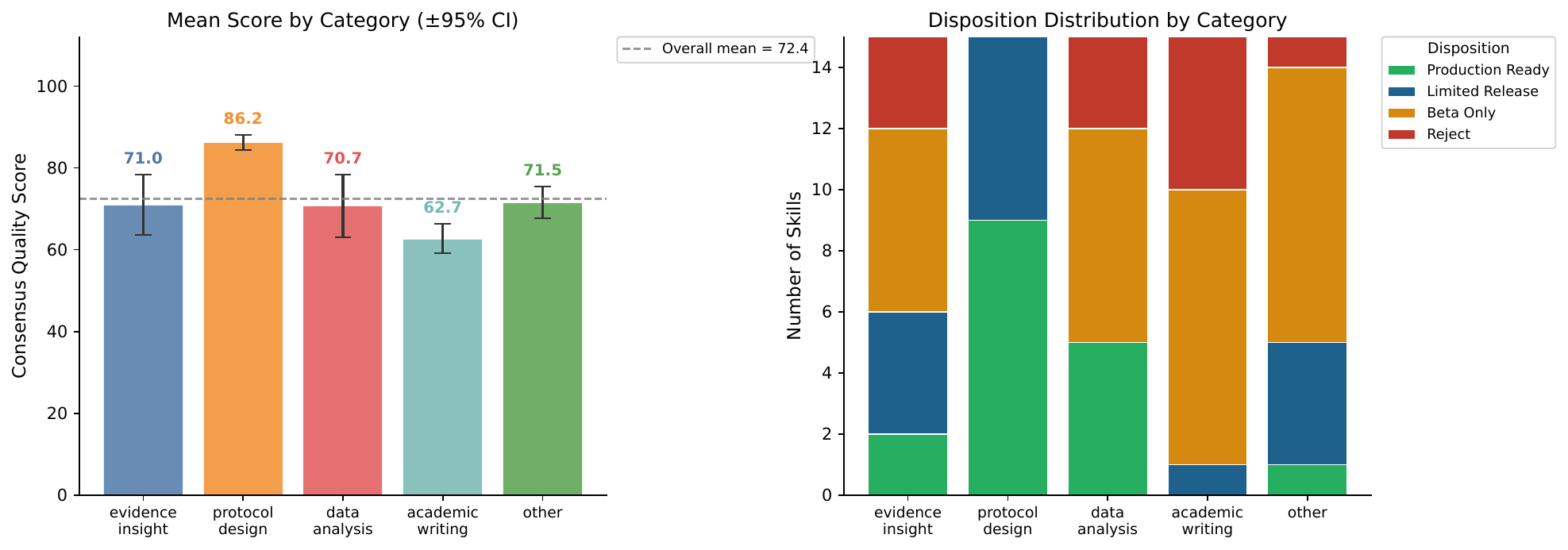}
\caption{Consensus quality scores by functional category. Left: mean $\pm$ 95\% CI per category; overall mean (72.4) shown as a dashed reference line. Right: stacked disposition bar chart showing the count of skills in each release tier by category.}
\end{figure}

The overall rate of expert disagreement requiring adjudication was 64.0\% (48/75). Adjudication was required for all 15 Academic Writing skills (100\%) and for 13/15 skills in the Other category (86.7\%), contrasting sharply with Protocol Design, where adjudication was required for only 1/15 skills (6.7\%).

\begin{table}[H]
\centering
\caption{Consensus quality score distribution by category (Experiment 1).}
\scriptsize
\resizebox{\textwidth}{!}{%
\begin{tabular}{l c c c c c}
\toprule
Category & n & Mean (SD) & Median & IQR & Adj. Rate \\
\midrule
Evidence Insight & 15 & 71.0 (14.7) & 73.2 & 63.2--83.3 & 60.0\% \\
Protocol Design & 15 & 86.2 (3.8) & 86.6 & 83.8--89.4 & 6.7\% \\
Data Analysis & 15 & 70.7 (15.3) & 71.0 & 63.6--86.9 & 66.7\% \\
Academic Writing & 15 & 62.7 (7.2) & 64.3 & 56.6--67.7 & 100.0\% \\
Other & 15 & 71.5 (7.8) & 74.2 & 67.0--76.9 & 86.7\% \\
Overall & 75 & 72.4 (13.0) & 73.2 & 64.4--84.1 & 64.0\% \\
\bottomrule
\end{tabular}%
}
\end{table}

\begin{figure}[H]
\centering
\includegraphics[width=0.92\textwidth]{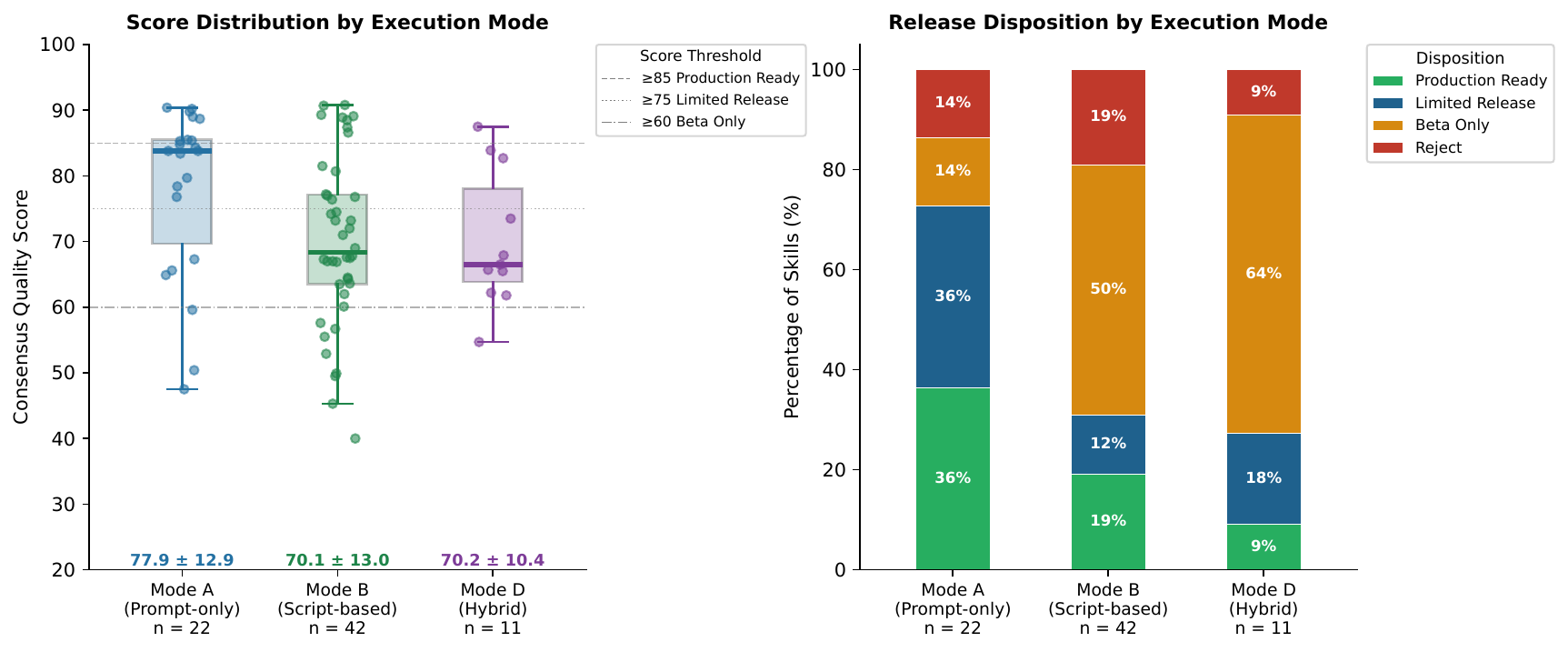}
\caption{Consensus quality score by execution mode (n = 75). Left: box and strip plot showing the score distribution for Mode A (prompt-only; n = 22), Mode B (script-based; n = 42), and Mode D (hybrid; n = 11); mean $\pm$ SD (ddof = 1) labelled below each group; horizontal reference lines indicate release disposition thresholds (dashed $\geq$ 85 Production Ready; dotted $\geq$ 75 Limited Release; dash-dot $\geq$ 60 Beta Only). Right: stacked proportion bar chart showing the release disposition breakdown for each execution mode. Mode A showed a higher mean consensus score (77.9 $\pm$ 12.9) and a greater proportion of Production Ready or Limited Release skills compared with Modes B (70.1 $\pm$ 13.0) and D (70.2 $\pm$ 10.4).}
\end{figure}

Applying the pre-specified optimization criteria defined in Section 2.6, 56 of 75 skills (74.7\%) were flagged for at least one optimization need. Twelve skills received a consensus Reject disposition (mean score = 52.0 $\pm$ 6.2; range: 40.0--59.6; Table 5). Of the 12 Reject skills, 8 also carried a high-risk flag (Y or Unclear) and all 12 required adjudication, indicating that even their failure status was contested between raters. An additional 9 skills received Beta Only dispositions with scores below 65 (mean = 63.4 $\pm$ 2.5), representing marginal deployability. Twenty-four skills carried a high-risk flag (Y or Unclear), spanning all five categories but concentrated in Data Analysis (n = 9) and Evidence Insight (n = 7).

\textit{Table 5. Skills with consensus Reject disposition, ordered by consensus quality score.}

\subsubsection{Representative Reject Cases}

The lowest-scoring consensus Reject skills are summarized below.

\begin{itemize}
\item \textbf{S009} (funding-trend-forecaster; Evidence Insight; 40.0; high-risk: Y): Mock data returned as real API results (M1, M4 FAIL).
\item \textbf{S031} (go-kegg-enrichment; Data Analysis; 45.3; high-risk: Y): Wrong function API; species mismatch in gene annotation (M3, M4 FAIL).
\item \textbf{S043} (meta-results-sensitivity-analysis; Data Analysis; 47.5; high-risk: Unclear): Scripts directory empty; declared functions unimplemented (T2, M4 FAIL).
\item \textbf{S003} (grant-funding-scout; Evidence Insight; 49.5; high-risk: Unclear): Hardcoded mock data undisclosed in SKILL.md (M1 FAIL).
\item \textbf{S054} (sci-paper-reviewer; Academic Writing; 49.9; high-risk: Unclear): Systematic dynamic evaluation failures across all test inputs.
\item \textbf{S039} (histolab; Data Analysis; 50.4; high-risk: Y): Critical dependency conflicts; cannot install (T1 FAIL).
\item \textbf{S053} (academic-highlight-generator; Academic Writing; 52.9; high-risk: N): Expert disagreement; shallow output quality.
\item \textbf{S055} (meta-manuscript-generator; Academic Writing; 54.7; high-risk: N): Structural generation failures under varied input conditions.
\item \textbf{S058} (biotech-pitch-deck-narrative; Academic Writing; 55.5; high-risk: N): Core logic incomplete; missing section scaffolding.
\item \textbf{S074} (kol-profiler; Other; 56.7; high-risk: N): Fundamental design gaps in core capability specification.
\item \textbf{S057} (microbiome-diversity-reporter; Academic Writing; 57.6; high-risk: Unclear): Result Determinism FAIL; outputs not reproducible (T3 FAIL).
\item \textbf{S010} (bibliography; Evidence Insight; 59.6; high-risk: Unclear): No executable code; all four Structural Veto dimensions FAIL.
\end{itemize}

\subsubsection{Agreement Analysis}

\textit{Human Inter-Rater Agreement (Baseline)}

Expert 1 and Expert 2 score agreement yielded an ICC(2,1) of 0.300 (95\% CI: 0.080--0.490; n = 74 complete pairs). Linearly weighted Cohen's $\kappa$ for disposition rank was 0.270. The mean absolute inter-expert score difference was 13.8 points (SD = 12.4; median = 9.0; maximum = 50.6). Exact rank agreement was achieved for 28/75 skills (37.3\%), and within-one-rank agreement for 56/75 (74.7\%). Under the Koo and Li framework, the observed ICC falls in the poor-to-fair reliability range, consistent with the high adjudication rate and reflecting the inherent subjectivity of evaluating complex multi-capability skills absent pre-calibration anchors [21]. The moderate human inter-rater baseline also establishes a realistic reference ceiling for automated agreement: a system that consistently approaches or exceeds this level of divergence magnitude can be considered a credible surrogate rater for first-pass governance purposes.

\textit{System--Expert Agreement}

The mean system quality score across all 75 skills was 71.0 (SD = 15.2), compared with a consensus mean of 72.4 (SD = 13.0). The mean system--consensus score difference ($\Delta$ = system - consensus) was -1.4 (SD = 14.8; range: -42.7 to +28.6), with no statistically significant directional bias (Wilcoxon W = 1258, p = 0.613).

\begin{figure}[H]
\centering
\includegraphics[width=0.92\textwidth]{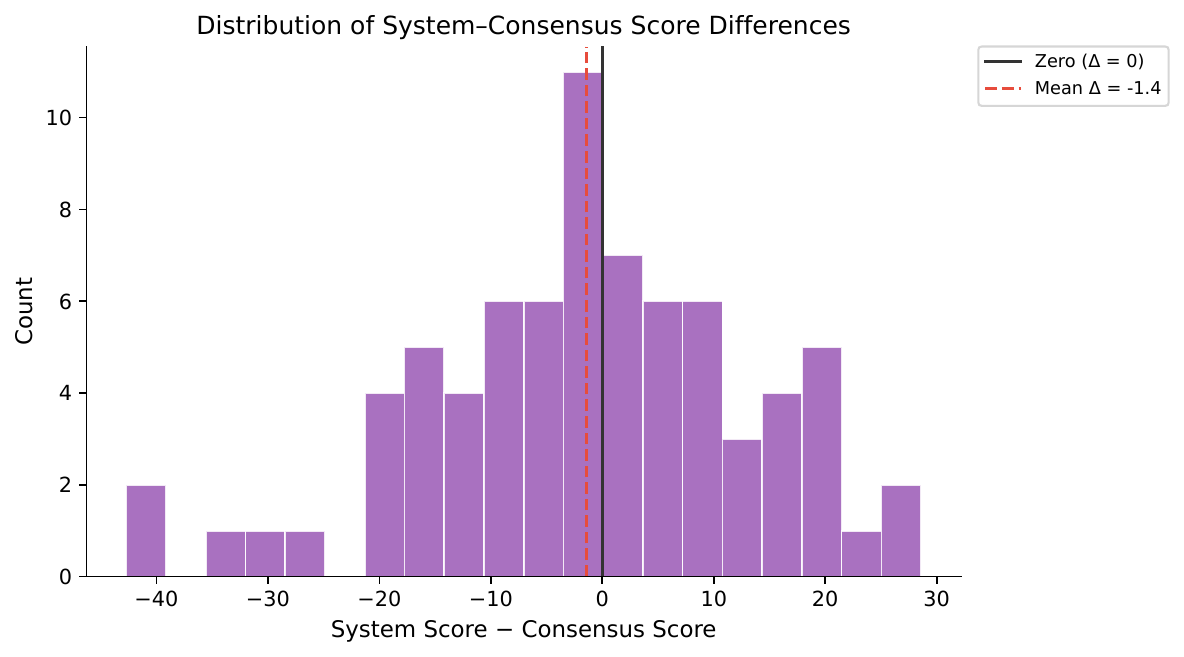}
\caption{Distribution of system--consensus score differences ($\Delta$ = system score - consensus score; n = 75). Vertical dashed line at $\Delta$ = 0; mean bias (-1.4) indicated. Positive values indicate system overscoring; negative values indicate system underscoring relative to expert consensus.}
\end{figure}

ICC(2,1) for system versus consensus was 0.449 (95\% CI: 0.250--0.610), exceeding the human inter-rater ICC of 0.300. Linearly weighted $\kappa$ for disposition rank was 0.215. The SD of absolute system--consensus differences (9.5) was lower than the SD of absolute inter-expert differences (12.4), indicating that the system's disagreement with expert consensus was no larger in magnitude than one expert's natural disagreement with the other. Exact system--consensus rank agreement was achieved for 22/75 skills (29.3\%), and within-one-rank agreement for 62/75 (82.7\%), exceeding the human within-one-rank rate of 74.7\%.

Bland-Altman analysis revealed a near-zero mean bias (-1.4) with limits of agreement of -29.0 to +26.2 (Figure 6). Two divergence patterns were identifiable. First, a subset of skills showed large negative $\Delta$ values (system $\ll$ consensus) arising from veto-driven score collapse: when a veto gate fired, the system score fell to the Reject range regardless of design quality, while experts --- evaluating under controlled conditions --- assigned substantially higher scores. Second, a smaller subset showed positive $\Delta$ values, reflecting the system's tendency to reward structural completeness while experts penalized shallow logic or limited use-case coverage.

\begin{figure}[H]
\centering
\includegraphics[width=0.92\textwidth]{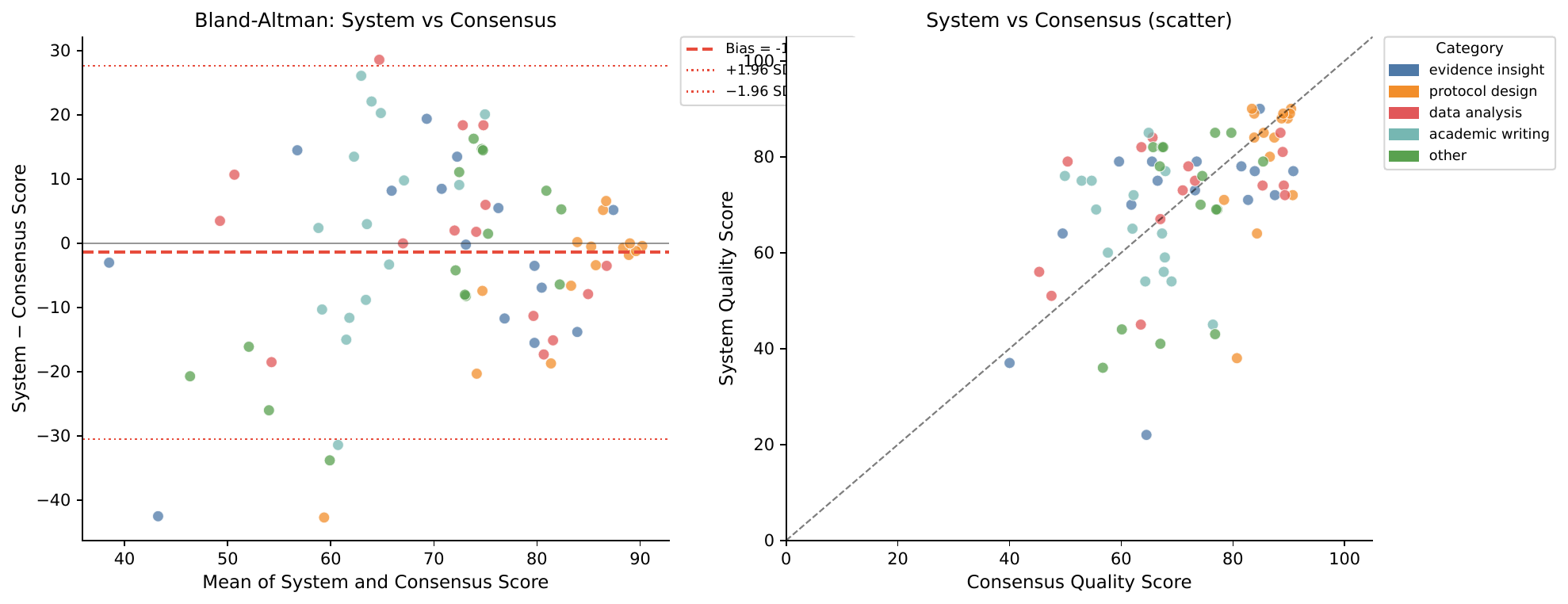}
\caption{Bland-Altman agreement plot (left) and scatter plot (right) for system vs. consensus quality scores. The solid line represents mean bias (-1.4); dashed lines represent 95\% limits of agreement (-29.0 to +26.2). The scatter plot diagonal represents perfect agreement.}
\end{figure}

\begin{table}[H]
\centering
\caption{Agreement summary for expert--expert and system--consensus comparisons.}
\scriptsize
\resizebox{\textwidth}{!}{%
\begin{tabular}{p{0.21\textwidth} c c c p{0.18\textwidth} c c}
\toprule
Comparison & n & Mean Bias & SD (diff) & ICC(2,1) [95\% CI] & Weighted $\kappa$ & Wilcoxon p \\
\midrule
Expert 1 vs. Expert 2 (baseline) & 74 & --- & 12.4 & 0.300 [0.080, 0.490] & 0.270 & --- \\
System vs. Consensus & 75 & -1.4 & 9.5 & 0.449 [0.250, 0.610] & 0.215 & 0.613 \\
\bottomrule
\end{tabular}%
}
\end{table}

\begin{figure}[H]
\centering
\includegraphics[width=0.92\textwidth]{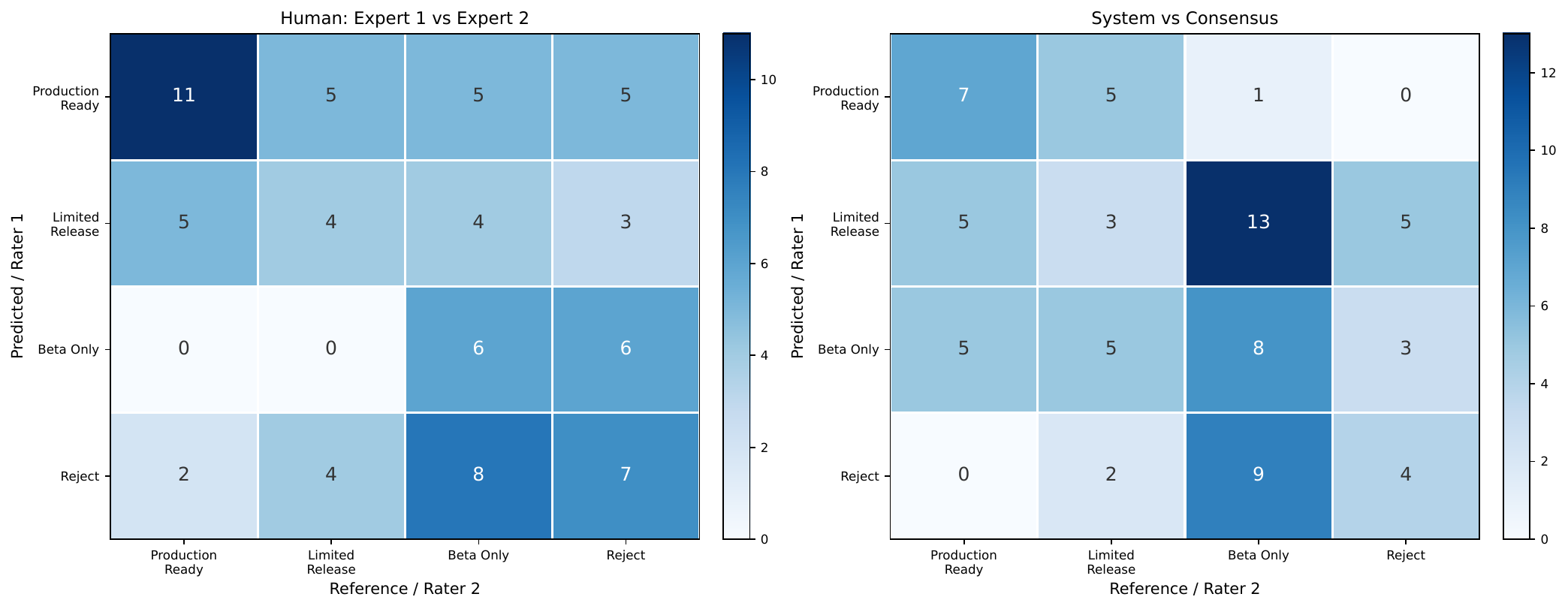}
\caption{Confusion matrices for release disposition rank agreement. Left: Expert 1 vs. Expert 2 (human baseline; exact agreement 37.3\%, within-1-rank 74.7\%). Right: System vs. Consensus (exact agreement 29.3\%, within-1-rank 82.7\%). Rows = actual disposition; columns = predicted/compared disposition.}
\end{figure}

\subsubsection{Stratified Agreement by Category}

Stratified ICC(2,1) values varied substantially across categories (Table 4; Figure 8). Evidence Insight and Data Analysis showed the highest ICC values (0.551 and 0.506, respectively), suggesting moderate agreement within those domains. Protocol Design showed lower ICC (0.232) and a statistically significant negative bias (mean $\Delta$ = -6.1; Wilcoxon p = 0.033), reflecting the system's tendency to apply Operational Stability veto gates to prompt-only protocol generation tools that human experts rated highly.

Academic Writing showed a pronounced negative ICC (-0.567; 95\% CI: -0.890 to -0.040) and a negative weighted $\kappa$ (-0.308), meaning that system and consensus scores moved in opposite directions within this category. When experts assigned a skill a high quality score, the system tended to score it lower, and vice versa. This pattern is attributable to a structural mismatch between the automated rubric and expert quality criteria for writing-focused skills: the system's Academic Tone dimension penalizes systematic paragraph structure and standard scientific hedging vocabulary as AI-stylistic markers, whereas expert raters regarded these features as indicators of clear and professionally appropriate scientific writing. The system's generic "Efficiency" criterion further disadvantages detailed academic outputs whose length reflects target deliverable standards rather than redundancy. No category-specific rubric override analogous to those implemented for Data Analysis and Protocol Design was available for Academic Writing in this version of the framework.

\begin{figure}[H]
\centering
\includegraphics[width=0.92\textwidth]{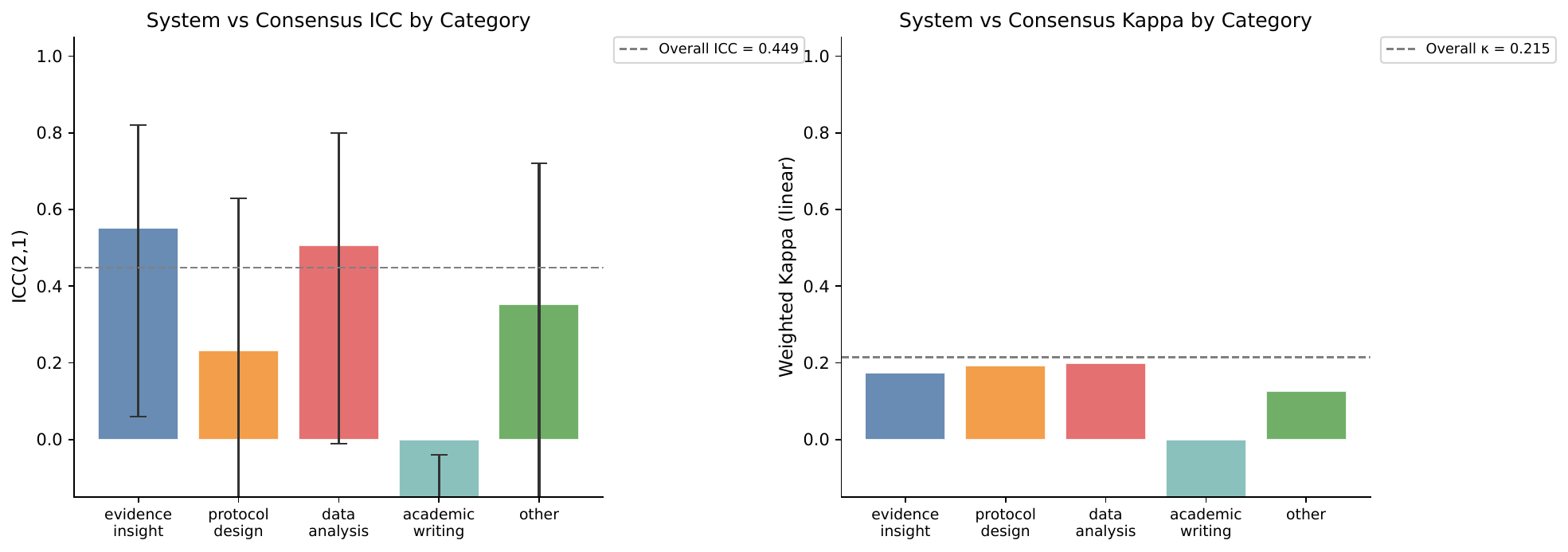}
\caption{Stratified ICC(2,1) and weighted $\kappa$ by functional category. Error bars represent 95\% confidence intervals. The dashed horizontal line at ICC = 0 is shown for reference; negative values indicate inverse agreement.}
\end{figure}

\begin{table}[H]
\centering
\caption{Stratified system--consensus agreement by functional category.}
\scriptsize
\resizebox{\textwidth}{!}{%
\begin{tabular}{p{0.21\textwidth} c c c p{0.18\textwidth} c c}
\toprule
Category & n & Mean Bias ($\Delta$) & SD & ICC(2,1) [95\% CI] & Weighted $\kappa$ & Wilcoxon p \\
\midrule
Evidence Insight & 15 & -1.5 & 15.5 & 0.551 [0.060, 0.820] & 0.174 & 1.000 \\
Protocol Design & 15 & -6.1 & 12.6 & 0.232 [-0.210, 0.630] & 0.194 & 0.033 \\
Data Analysis & 15 & +1.1 & 13.9 & 0.506 [-0.010, 0.800] & 0.198 & 0.851 \\
Academic Writing & 15 & +3.1 & 16.4 & -0.567 [-0.890, -0.040] & -0.308 & 0.489 \\
Other & 15 & -3.5 & 15.7 & 0.353 [-0.170, 0.720] & 0.126 & 0.561 \\
\bottomrule
\end{tabular}%
}
\end{table}

\begin{figure}[H]
\centering
\includegraphics[width=0.92\textwidth]{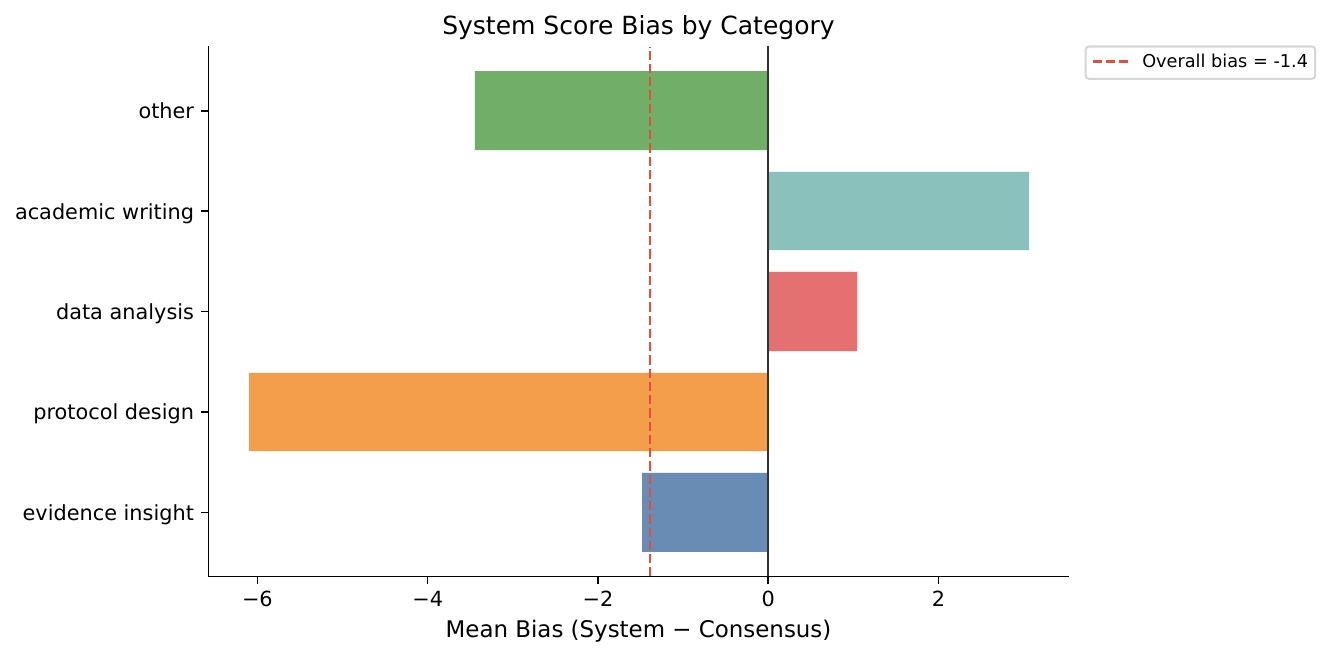}
\caption{Mean system--consensus score bias by category. Error bars represent $\pm$1 SD. The asterisk (*) on Protocol Design indicates a statistically significant bias (Wilcoxon p = 0.033). Positive bias indicates system overscoring; negative bias indicates system underscoring.}
\end{figure}

\section{Discussion}

\subsection{A Structured Audit Framework Addresses a Distinct Gap in Medical AI Skill Governance}

The central finding of this study is that structured, domain-specific automated audit can achieve agreement with expert review that, in terms of score divergence magnitude, falls within the natural range of human inter-rater variability. This result is notable for two reasons. First, it demonstrates that the construct of skill release readiness --- which encompasses structural integrity, scientific validity, dynamic execution behavior, and safety boundaries --- can be operationalized sufficiently to produce assessments that are not systematically more divergent than those produced by a second human rater. Second, it provides a principled basis for deploying MedSkillAudit as a first-pass governance layer. Of the 12 Reject-disposition skills, 7 were flagged via hard veto gates (Structural or Research); the remaining 5 fell below the 60-point threshold through rubric penalties on dynamic evaluation. All 12 received dispositions that would prevent deployment, demonstrating that the framework identifies catastrophic failures across both gate-triggered and score-based failure pathways, without generating the volume of false positives that would undermine its practical utility as a screening tool.

This positions MedSkillAudit distinctly from existing approaches to evaluating medical AI systems. Prior work has focused primarily on benchmark performance of foundation models [6--8] or agentic systems in clinical task environments [4, 11]. These approaches assess what a model \textit{can do} on a curated test set, but they do not evaluate whether a skill \textit{should be deployed} as a reusable artifact in a real research workflow --- a question involving source code quality, dependency reliability, output reproducibility, and scientific integrity of generated content [3, 5, 12]. MedSkillAudit addresses this deployment-readiness dimension directly, filling a governance gap that neither static capability benchmarks nor general-purpose code quality tools were designed to address.

\subsection{Architectural Strengths: Hard Gates, Dynamic Execution, and Multi-Layer Scoring}

Several design choices in MedSkillAudit contributed to its reliability against expert review. The two-gate veto architecture --- separating structural integrity (Veto Gate 1) from scientific integrity (Veto Gate 2) --- reflects the distinct risk profiles of different failure modes. A skill that crashes on 30\% of inputs or contains unsanitized exec() calls poses a deployment risk that is categorically different from a skill with mediocre but complete outputs; treating these as point deductions on a continuous scale would obscure their qualitative difference. The binary veto design ensures that these failure classes cannot be numerically compensated by high scores elsewhere --- a property that proved empirically important, as the Research Veto's Code Usability gate (M4) correctly identified six skills whose documentation appeared complete but whose generated code was non-functional, a class of failure that documentation-only static review would not detect.

The 60\% weight on dynamic execution testing is similarly well-grounded. For the 53 script-based or hybrid skills in the evaluation set (Modes B and D), static analysis of a SKILL.md cannot detect whether a declared function exists in the installed library version, whether a bioinformatics API has changed, or whether a dependency conflict prevents installation entirely. Dynamic execution exposed these failures directly and consistently. The mean dynamic score correlation with consensus was stronger than the static score correlation across all three technical categories (Evidence Insight, Data Analysis, Protocol Design), supporting the rationale for the asymmetric 40/60 weighting.

The two-layer specialized rubric architecture contributes differentiated signal that generic quality rubrics cannot provide. The Category 3 Code Executability sub-dimension, for instance, provides a graduated score that captures partially functional code --- useful diagnostic information even when Veto Gate 2 has already flagged the skill for rejection. The Category 2 Evidence Hierarchy Matching and Validation dimensions captured methodologically inconsistent protocols that passed all structural checks. These category-specific rubric layers transform the audit from a pass/fail screener into a structured diagnostic that supports targeted revision.

The eight-dimensional ISO 25010-aligned static scoring system, covering functional suitability, reliability, usability, performance efficiency, maintainability, portability, security, and compatibility, ensures that the audit surface is comprehensive and grounded in an established quality standard rather than ad hoc criteria. This is particularly valuable in a pre-deployment governance context, where the audit must be defensible to stakeholders with varied technical backgrounds.

\subsection{Framework Calibration Is an Iterative Process: Evidence from v1.0 to v1.1.0}

An important secondary finding of this study is that the framework's limitations --- where they exist --- are diagnosable and addressable through targeted rubric adjustments, rather than reflecting fundamental architectural flaws. Based on the diagnostic evidence from Experiment 1, we iteratively refined the framework to skill-auditor@1.1.0 after data collection was complete. The following describes the calibration rationale and the specific rubric changes incorporated into v1.1.0. These refinements have not yet been evaluated in a controlled experiment; their expected impact on category-level reliability is discussed in Section 4.6.

During the audit of a reference Data Analysis skill (\textit{differential-expression-analysis}), three systematic biases in the shared basic evaluation rubric were identified: (1) the Fault Tolerance criterion (Basic Evaluation 2.1) penalizes hard stops on invalid inputs as failures of graceful degradation, whereas in scientific computing pipelines, halting at the source of a malformed sample ID is the correct defensive behavior that prevents silent data corruption; (2) the Forgiveness criterion (Basic Evaluation 5.2) rewards auto-correction of ambiguous inputs, whereas in statistical analysis, fuzzy-matching parameter inputs risks producing biologically incorrect results without warning; and (3) the Recoverability criterion (Basic Evaluation 2.3) rewards inline human-readable recovery guidance, whereas agent-first execution contexts require structured machine-parseable error codes rather than terminal prose. All three biases arise from the shared rubric's implicit assumption of human-operated general-purpose software --- an assumption that does not hold for scientific computing skills operating in agent-first Mode B/D contexts.

Rather than modifying the shared rubric and affecting all five categories, version 1.1.0 introduced per-category Scene Override sections in the specialized evaluation files for Categories 2 (Protocol Design) and 3 (Data Analysis), with an Execution Mode Awareness note added for Category 5 (Other) when operating in agent-first contexts. This approach preserves the integrity of the shared rubric for categories where its assumptions hold, while correcting the systematic misclassifications for scientific computing skills. The Protocol Design negative bias observed in Experiment 1 (mean $\Delta$ = -6.1, p = 0.033) is the expected empirical signature of exactly this class of misclassification --- and is the target of the Category 2 override in v1.1.0. This iterative calibration pathway --- audit a reference skill, identify systematic rubric mismatches, add targeted scene overrides --- represents a principled and sustainable mechanism for improving framework reliability across categories over successive versions.

\subsection{The Academic Writing ICC as a Diagnostic, Not a Failure}

The pronounced negative ICC for Academic Writing (-0.567) warrants careful interpretation. A negative ICC does not indicate random disagreement; it indicates that the system and expert consensus are moving in \textit{opposite directions} --- when experts rate a writing skill highly, the system tends to score it lower, and vice versa. This is a structural diagnostic that reveals a specific mismatch between what the automated rubric measures and what expert raters value in this category.

A key factor underlying this divergence is that expert evaluators of writing-focused skills orient primarily toward the quality of the generated artifact --- whether the output is scientifically accurate, structurally complete, appropriately hedged, and professionally registered --- rather than toward the operational characteristics of the AI process that produced it. Experts do not assess whether the skill's internal prompting is efficient, whether its output style pattern-matches to AI-generation heuristics, or whether the skill terminates quickly on edge-case inputs. The automated rubric, designed with the same assumptions as its technical counterparts, applies an Efficiency criterion that penalizes detailed academic outputs for length, and an Academic Tone dimension that treats systematic paragraph structure and standard hedging vocabulary as AI-stylistic markers rather than scientific writing norms. The result is that the system and experts are, to a significant degree, measuring different constructs within this category.

Crucially, this divergence is not uninformative --- it is precisely what a domain-specific audit framework \textit{should} reveal. The fact that expert raters assign high scores to Academic Writing skills while the automated system surfaces structural concerns (incomplete section coverage, non-deterministic outputs, absent disclaimer statements) suggests that expert evaluation in this category may be anchored primarily on the best-case output quality rather than on the systematic behavior of the skill under varied inputs. MedSkillAudit's strength is that it probes the full behavioral distribution across multiple test inputs, not just the quality of a single representative output. The negative ICC therefore points to a class of risk that expert spot-checking is poorly positioned to detect: Academic Writing skills that produce impressive outputs in typical conditions but exhibit reproducibility failures, incomplete section generation under adversarial prompts, or missing scientific integrity disclaimers when input parameters shift. A well-designed audit rubric for Category 4 should synthesize both dimensions --- expert-style artifact quality assessment and system-style behavioral reliability testing --- rather than privileging either alone.

\subsection{Practical Implications for Skill Governance}

MedSkillAudit is designed for integration into pre-deployment workflows at research institutions managing skill libraries. The framework's layered veto architecture enables early identification of catastrophic failures --- skills that are unsafe to deploy regardless of design quality --- while the structured scoring system supports prioritized revision efforts for borderline cases. Research teams can apply MedSkillAudit iteratively: initial audit identifies high-risk skills and surfaces per-dimension score gaps, targeted refinement addresses the specific rubric dimensions flagged in the audit report, and re-audit confirms quality improvements before deployment. The framework's disposition tiers provide a transparent and defensible basis for governance decisions: skills in the Beta Only tier can be staged for limited testing environments while revision proceeds, and the structured JSON output from each audit run creates an auditable record of quality state over time.

The scene override mechanism provides a practical pathway for institutions to adapt MedSkillAudit to domain-specific use cases without redesigning the core pipeline. Institutions operating in specialized subfields --- such as clinical trial design, regulatory submission drafting, or patient-facing communication tools --- can introduce targeted rubric adjustments that reflect their operational constraints, while retaining the shared structural and scientific integrity gates that apply across all medical research contexts.

\subsection{Limitations and Directions for Future Work}

This study has several limitations that should be acknowledged. First, the evaluation set of 75 skills, while sufficient for a preliminary reliability study, limits the statistical power of stratified analyses (n = 15 per category), and the per-category ICC confidence intervals are correspondingly wide. Larger evaluation sets per category would allow more precise estimation of category-specific agreement and would better characterize the tails of the system--consensus divergence distribution.

Second, the human inter-rater ICC of 0.300 --- a relatively modest baseline --- reflects the absence of pre-rating calibration sessions for the expert reviewers in Experiment 1. The 64\% adjudication rate and 100\% Academic Writing adjudication rate indicate that the evaluation rubric, while comprehensive, would benefit from annotated tier-anchor examples that reduce evaluator interpretation variability. Planned calibration rounds ahead of future experiments will address this directly.

Third, the framework's current rubric weighting (40\% static, 60\% dynamic) is fixed across execution modes, creating a known structural mismatch for Mode A prompt-only skills, where dynamic evaluation measures a running Claude agent rather than a fixed code artifact. A mode-adaptive weighting scheme --- increasing the static weight for Mode A skills where the SKILL.md design is the primary artifact --- would better reflect the quality signals available in different execution contexts. Similarly, the Category 4 rubric requires a targeted scene override analogous to those developed for Categories 2 and 3, specifically recalibrating the Efficiency and Academic Tone dimensions for writing-focused skills.

Looking ahead, MedSkillAudit is designed as a living framework whose rubric can be iteratively refined as new skill categories and execution patterns emerge. Unlike Categories 2 and 3, the Academic Writing (Category 4) scene override has not yet been implemented in any released version of the framework; targeted rubric recalibration for the Efficiency and Academic Tone dimensions is the primary planned change for skill-auditor@1.2.0, alongside a mode-adaptive static/dynamic weighting scheme for Mode A skills. These changes are expected to reduce the Category 4 ICC inversion and the Mode A score inflation observed in Experiment 1, but will require a controlled re-evaluation to confirm. More broadly, the scene override mechanism provides a principled pathway for extending the framework to new medical research domains --- such as clinical trial design, regulatory submission drafting, or patient-facing communication tools --- without requiring a redesign of the core audit architecture. Pre-deployment audit frameworks of this kind may become increasingly important as medical research institutions begin to govern skill libraries as production infrastructure rather than experimental tools.

\section{Data and Code Availability Statement}

The 75 medical research agent skills comprising the evaluation set were drawn from historical development versions produced across two independent research teams. Due to institutional data governance restrictions, the historical version data of these skills is not publicly available. Aggregated quality metrics, disposition counts, and anonymized per-category agreement statistics are reported within the article. MedSkillAudit (skill-auditor) is implemented as a medical research agent skill; the current version, including the full audit pipeline, Veto Gate criteria, the Layer 1 generic rubric, and the five category-specific Layer 2 rubrics, is publicly available at \url{https://github.com/aipoch/medical-research-skills} under the MIT license, along with the latest versions of these 75 medical research agent skills.

\typeout{get arXiv to do 4 passes: Label(s) may have changed. Rerun}
\end{document}